\documentclass{svproc}
\usepackage[utf8]{inputenc}
\usepackage{graphicx}
\usepackage{amssymb,xcolor}
\usepackage{cancel}
\usepackage{enumitem}
\usepackage[noend]{algorithmic}
\usepackage{algorithm}
\usepackage{wrapfig}
\usepackage{hyperref}

\newcommand\blfootnote[1]{%
  \begingroup
  \renewcommand\thefootnote{}\footnote{#1}%
  \addtocounter{footnote}{-1}%
  \endgroup
}
\usepackage[colorinlistoftodos]{todonotes}
\usepackage{subcaption}

\title{Information Requirements of Collision-Based Micromanipulation} 
\author{Alexandra Q. Nilles$^{*,1}$, Ana Pervan$^{*,2}$, Thomas A. Berrueta$^{*,2}$,\\ Todd D. Murphey$^2$ \and Steven M. LaValle$^{1,3}$}
\authorrunning{Nilles, Pervan, Berrueta, \textit{et al.}}
\institute{$^1$Department of Computer Science, \\ University of Illinois at Urbana-Champaign, Urbana, IL, USA \\ 
$^2$Department of Mechanical Engineering,\\ Northwestern University, Evanston, IL, USA \\ 
$^3$Faculty of Information Technology and Electrical Engineering,
\\University of Oulu, Oulu, Finland}

\begin{document}

\maketitle
\vspace{-0.25in}
\begin{abstract}
We present a task-centered formal analysis of the relative power of several robot designs, inspired by the unique properties and constraints of micro-scale robotic systems. Our task of interest is object manipulation because it is a fundamental prerequisite for more complex applications such as micro-scale assembly or cell manipulation. Motivated by the difficulty in observing and controlling agents at the micro-scale, we focus on the design of \emph{boundary interactions}: the robot's motion strategy when it collides with objects or the environment boundary, otherwise known as a \emph{bounce rule}. We present minimal conditions on the sensing, memory, and actuation requirements of periodic ``bouncing'' robot trajectories that move an object in a desired direction through the incidental forces arising from robot-object collisions. Using an information space framework and a hierarchical controller, we compare several robot designs, emphasizing the information requirements of goal completion under different initial conditions, as well as what is required to recognize irreparable task failure. 
Finally, we present a physically-motivated model of boundary interactions, and analyze the robustness and dynamical properties of resulting trajectories.

\end{abstract}

\vspace{-0.35in}
\section{Introduction}
\vspace{-0.2in}

\label{sec-intro}
\blfootnote{$^*$These authors contributed equally to the work. \\ e-mails: \href{mailto:nilles2@illinois.edu}{\nolinkurl{nilles2@illinois.edu}}, \href{mailto:anapervan@u.northwestern.edu}{\nolinkurl{anapervan@u.northwestern.edu}}, \href{mailto:tberrueta@u.northwestern.edu}{\nolinkurl{tberrueta@u.northwestern.edu}}, \href{mailto:t-murphey@northwestern.edu}{\nolinkurl{t-murphey@northwestern.edu}}, \href{mailto:steven.lavalle@oulu.fi}{\nolinkurl{steven.lavalle@oulu.fi}}}

Robots at the micro-scale have unique constraints on the amount of possible onboard information processing. Despite this limitation, future biomedical applications of micro-robots, such as drug delivery, tissue grafting, and minimally invasive surgery, demand sophisticated locomotion, planning, and manipulation~\cite{Sitti2015,Soto2018}. While certain robotic systems have succeeded at these tasks with assistance from external sensors and actuators~\cite{Xu2015}, the minimal sensing and actuation requirements of these tasks are not well-understood.

Ideally, designs at this scale would not require fine-grained, individual motion control, due to the extreme difficulty in observing and communicating with individual agents. In fact, micro-scale locomotion is often a direct consequence of the fixed or low degree-of-freedom morphology of the robot, suggesting that direct co-design of robots and their motion strategies may be necessary~\cite{Censi2016}. The work presented in this manuscript provides the beginning of a theory of task-centered robot design, used to devise micro-robot morphology and propulsion mechanisms. In order to inform the design of task-capable micro-robotic platforms, we analyze the information requirements of tasks~\cite{donald1997}.

We focus on the task of {\em micromanipulation} with micro-robots, a fundamental task underlying more complex procedures such as drug delivery and cell transplantation~\cite{Li2017}. In this paper, we investigate micro-robot motion strategies that explicitly use {\em boundary interactions}: the robot's action when it encounters an environment boundary.  Identifying minimal information requirements in this setting is essential for reasoning about robot performance, and is also a first step toward automating co-design of robots and their policies. In Section \ref{sec-modeldefs}, we define abstract yet physically motivated models, aiming to roughly cover an interesting set of possible micro-robot realizations. Section \ref{sec-motivation} motivates our task of interest and our assumptions. In Section \ref{sec-designs}, we compare several robot designs and state results on requirements for task completion, with respect to controller complexity, sensor power, and onboard memory. Finally in Section \ref{sec-dyn} we state results for the robustness of the periodic trajectories used in the hierarchical controller.

\vspace{-0.1in}
\subsection{Background and Related Work}

\label{sec-background}
The scientific potential of precise manipulation at microscopic scales has been appreciated for close to a century~\cite{Chambers1931}. As micro/nano-robots have become increasingly sophisticated, biomedical applications such as drug delivery~\cite{Li2017} and minimally-invasive surgery~\cite{Soto2018} have emerged as grand challenges in the field~\cite{Yang2018}.

To formally analyze the information requirements of planar micromanipulation, we deliberately abstract many physically-motivated actuators and sensors, such as odometry, range-sensing, differential-drive locomotion, and others. Additionally, similar to the work in~\cite{donald1997}, we avoid tool-specific manipulation by purposefully abstracting robot-object contacts in an effort to be more general. However, we focus on collision-based manipulation instead of push-based. Recent publications have illustrated the value of robot-boundary collisions in generating reliable and robust robot behaviors~\cite{kim2015,Nilles2018,Savoie2019}. Thus, through deliberate abstraction we develop and analyze a model of collision-based micromanipulation that can serve as a test-bed for micro-robotic designs. 

Despite substantial advances in micro-robotics, agents at micro/nano length-scales face fundamental difficulties and constraints. For one, as robots decrease in size and mass, common components such as motors, springs and latches experience trade-offs in force/velocity generation that limit their output and constrain the space of feasible mechanical designs~\cite{Ilton2018}. Furthermore, battery capacities and efficiencies~\cite{Oudenhoven2011}, as well as charging and power harvesting~\cite{Ding2013}, experience diminishing performance at small scales, which restrict electrical designs of micro-machines. These limitations collectively amount to a rebuke of traditionally complex robotic design at the length-scales of interest, and suggest that a minimalist approach may be necessary. The minimal approach to robot design asks, ``what is the most simple robot that can accomplish a given task?''

Minimalist approaches to microrobot design often exploit the close relationship between morphology and computation. For example, a DNA-based nano-robot is capable of capturing and releasing molecular payloads in response to a binary stimulus~\cite{Douglas2012}. The inclusion of a given set of sensors or actuators in a design can enable robotic agents to sidestep complex computation and planning in favor of direct observation or action. Alternatively, in top-down approaches the formulation of high-level control policies can guide the physical design of robots~\cite{Pervan2018}. While many approaches have seen success in their respective domains, the problem of optimizing the design of a robot subject to a given task has been shown to be NP-hard~\cite{Saberifar2018}. 

Due to information constraints, designing control policies for minimal robots remains a challenge. In order to achieve complex tasks, controllers are often hand-tuned to take advantage of the intrinsic dynamics of the system. For example, in~\cite{alam2017minimalist,Alam2018} the authors show that one can develop controllers for minimal agents (solely equipped with a clock and a contact sensor) that can achieve spatial navigation, coverage, and localization by taking advantage of the details of the agents' dynamics. Hence, in order to develop control strategies amenable to the constraints of such robots, one requires substantial analytical understanding of the capabilities of minimal robots. 

A unifying theory of robotic capabilities was established in~\cite{okane2006}. The authors develop an information-based approach to analyzing and comparing different robot designs with respect to a given task. The key insight lies in distilling the minimal information requirements for a given task and expressing them in an appropriately chosen information space for the task. Then, as long as we are capable of mapping the individual information histories of different robot designs into the task information space, the performance of robots may be compared. Our main contribution to this body of work is a novel demonstration of the approach in~\cite{okane2006} to an object manipulation task. Additionally, we use a hierarchical approach to combine the resulting high-level task completion guarantees with results on the robustness of low-level controllers.

\vspace{-0.1in}
\section{Model and Definitions}
\vspace{-0.05in}
\label{sec-modeldefs}
Here we introduce relevant abstractions for characterizing robots generally, as well as their capabilities for given tasks. We largely follow from the work of~\cite{LaValle2006,okane2006}.

\vspace{-0.05in}
\subsection{Primitives and Robots}
\vspace{-0.05in}
\label{sec-robotprims}
In this work, a robot is modelled as a point in the plane; this model has obvious limitations, but captures enough to be useful for many applications, especially \emph{in vitro} where the robot workspace is often a thin layer of fluid. Hence, its configuration space is $X \subseteq SE(2)$, and its configuration is represented as $(x,y,\theta)$. The robot's environment is $E\subseteq\mathbb{R}^2$, along with a collection of lines representing boundaries; these may be one-dimensional ``walls'' or bounded polygons. The environment may contain objects that will be static unless acted upon.

Following the convention of \cite{okane2006}, we define a robot through sets of primitives. A \emph{primitive}, $P_i$, defines a ``mode of operation" of a robot, and is a 4-tuple $P_i = (U_i,Y_i,f_i,h_i)$ where $U_i$ is the action set, $Y_i$ is the observation set, $f_i: X \times U_i \to X$ is the state transition function, and $h_i: X \times U_i \to Y_i$ is the observation function. Primitives may correspond to use of either a sensor or an actuator, or both if their use is simultaneous (see~\cite{LaValle2006} for examples). We will model time as proceeding in discrete stages. At stage $k$ the robot occupies a configuration $x_k\in X$, observes sensor reading $y_k^i\in Y_i$, and chooses its next action $u_k^i\in U_i$ for each $i$ primitive in its set. A \textit{robot} may then be defined as a 5-tuple $R=(X,U,Y,f,h)$ comprised of the robot's configuration space in conjunction with the elements of the primitives 4-tuples. With some abuse of notation, we occasionally write robot definitions as $R = \{P_1,\ ...,\ P_N\}$ when robots share the same configuration space to emphasize differences between robot capabilities.

\vspace{-0.05in}
\subsection{Information Spaces}
\vspace{-0.05in}
\label{sec-ispace}
A useful abstraction to reason about robot behavior in the proposed framework is the information space (I-space). Information spaces are defined according to actuation and sensing capabilities of robots, and depend closely on the robot's history of actions and measurements. We denote the {\em history} of actions and sensor observations at stage $k$ as $(u_1,y_1,\dots,u_{k},y_k)$. The history, combined with initial conditions $\eta_0=(u_0,y_0)$, yields the \emph{history information state}, $\eta_k = (\eta_0,u_1,y_1,\dots,u_{k},y_k)$. In this framework, initial conditions may either be the exact starting state of the system, or a set of possible starting states, or a prior belief distribution. The collection of all possible information histories is known as the \emph{history information space}, $\mathcal{I}_{hist}$. It is important to note that a robot's history I-space is intrinsically defined by the robot primitives and its initial conditions. Hence, it is not generally fruitful to compare the information histories of different robots.

{\em Derived} information spaces should be constructed to reason about the capabilities of different robot designs. A derived I-space is defined by an \textit{information map} $\kappa: \mathcal{I}_{hist} \to \mathcal{I}_{der}$ that maps histories in the history I-space to states in the derived I-space $\mathcal{I}_{der}$. Mapping different histories to the same derived I-space allows us to directly compare different robots. The exact structure of the derived I-space depends on the task of interest; an abstraction must be chosen that allows for both meaningful comparison of the robots as well as determination of task success.

In order to be able to compare robot trajectories within the derived I-space, we introduce an \textit{information preference relation} to distinguish between derived information states~\cite{okane2006}. We discriminate these information states based on a distance metric to a given goal region $\mathcal{I}_G\subseteq\mathcal{I}_{der}$ which represents success for a task. Using a relation of this type we can assess ``preference" over information states, notated as $\kappa(\eta^{(1)}) \preceq \kappa(\eta^{(2)})$ if an arbitrary $\eta^{(2)}$ is preferred over $\eta^{(1)}$. 

\vspace{-0.05in}
\subsection{Robot Dominance}
\vspace{-0.05in}
\label{sec-dominance}
Extending on the definition of information preference relations in the previous section, we can define a similar relation to compare robots. Here, we define a relation that captures a robot's ability to ``simulate" another given a policy. Policies are mappings $\pi$ from an I-space to an action set: the current information state determines the robot's next action. We additionally define a function $F$ that iteratively applies a policy to update an information history. The updated history I-state is given by $\eta_{m+k}=F^m(\eta_k,\pi,x_k)$, where $x_k\in X$.

\vspace{0.05in}
\noindent\textbf{Definition 1. (Robot dominance} from \cite{okane2006}\textbf{)} Consider two robots with a task specified by reaching a goal region $\mathcal{I}_G\subseteq\mathcal{I}_{der}$: 
\vspace{-0.1in}
\[R_1 = (X^{(1)},U^{(1)},Y^{(1)},f^{(1)},h^{(1)})\]
\vspace{-0.3in}
\[R_2 = (X^{(2)},U^{(2)},Y^{(2)},f^{(2)},h^{(2)}).\]
\vspace{-0.2in}

\noindent Given I-maps $\kappa_1:\mathcal{I}_{hist}^{(1)} \to \mathcal{I}_{der}$ and $\kappa_2:\mathcal{I}_{hist}^{(2)} \to \mathcal{I}_{der}$, if for all: $\eta^{(1)} \in \mathcal{I}_{hist}^{(1)}$ and $\eta^{(2)} \in \mathcal{I}_{hist}^{(2)}$ for which $\kappa_1(\eta^{(1)}) \preceq \kappa_2(\eta^{(2)})$; and $u^{(1)} \in U^{(1)}$; there exists a policy, defined as $\pi_2: \mathcal{I}_{hist}^{(2)} \to U^{(2)}$, generating actions for $R_2$ such that for all $x^{(1)} \in X^{(1)}$ consistent with $\eta^{(1)}$ and all $x^{(2)} \in X^{(2)}$ consistent with $\eta^{(2)}$, there exists a positive integer $l$ such that 
\vspace{-0.1in}
\[\kappa_1(\eta^{(1)},u^{(1)},h^{(1)}(x^{(1)},u^{(1)})) \preceq \kappa_2(F^l(\eta^{(2)},\pi_2,x^{(2)}))\]
\vspace{-0.2in}

\noindent then $R_2$ \emph{dominates} $R_1$ under $\kappa_1$ and $\kappa_2$, denoted $R_1 \unlhd R_2$. If both robots can simulate each other ($R_1 \unlhd R_2$ and $R_2 \unlhd R_1$), then $R_1$ and $R_2$ are \emph{equivalent}, denoted by $R_1 \equiv R_2$. Lastly, we introduce the following lemma:

\vspace{0.05in}
\noindent\textbf{Lemma 1.} (from \cite{okane2006}) Consider three robots $R_1$, $R_2$, and $R_3$ and an I-map $\kappa$. If $R_1 \unlhd R_2$ under $\kappa$, we have:
\vspace{-0.05in}
\begin{enumerate}[label=(\alph*)]
\item $R_1 \unlhd R_1 \cup R_3$ (adding primitives never hurts);
\item $R_2 \equiv R_2 \cup R_1$ (redundancy does not help);
\item $R_1 \cup R_3 \unlhd R_2 \cup R_3$ (no unexpected interactions).
\end{enumerate}

\vspace{-0.25in}

\begin{figure}[pt]
\centering
\includegraphics[width=0.95\linewidth]{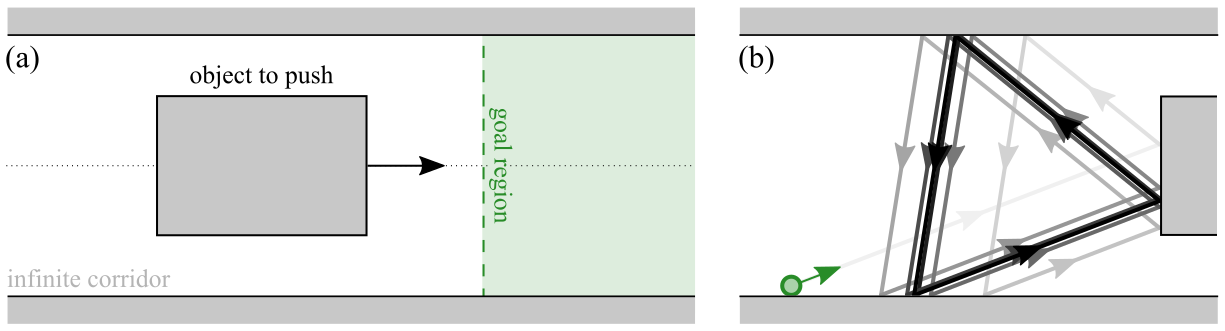}
\caption{\label{fig:env_tri} (a) A rectangular object in a long corridor that can only translate to the left or right, like a cart on a track. The robot's task is to move the object into the green goal region. (b) The robot, shown in green, executes a trajectory in which it rotates the same relative angle each time it collides.}\vspace{-0.2in}
\end{figure}

\vspace{-0.0in}
\section{Manipulating a Cart in a Long Corridor}
\vspace{-0.1in}
\label{sec-motivation}

We begin our analysis of the information requirements of micro-scale object manipulation by introducing a simple, yet rich, problem of interest. Consider a long corridor, containing a rectangular object, as shown in Fig.~\ref{fig:env_tri}(a). The object may only translate left or right down the corridor, and cannot translate toward the corridor walls or rotate. We can abstract this object as a cart on a track; physically, such one-dimensional motion may arise at the micro-scale due to electromagnetic forces from the walls of the corridor, from fluid effects, or from direct mechanical constraints. The task for the robot is then to manipulate the object into the goal region (green-shaded area in Fig.~\ref{fig:env_tri}(a)). We believe that this simplified example will illustrate several interesting trade-offs in the sensing, memory, and control specification complexity necessary to solve the problem. On its own merits, this task solves the problem of \emph{directed transport}, and can be employed as a useful component in larger, more complex microrobotic systems.

To break the symmetry of the problem, we assume that the object has at least two distinguishable sides (left and right). For example, the two ends of the object may emit chemical A from the left side and chemical B from the right side. The robot may be equipped with a chemical comparator that indicates whether the robot is closer to source A or source B (with reasonable assumptions on diffusion rates). The object may have a detectable, directional electromagnetic field. All these possible sensing modalities are admissible under our model. We also assume that the dimensions of the corridor and the object are known and will be used to design the motion strategy of the robot---often a fair assumption in laboratory micro-robotics settings.

Here, we investigate the requirements of {\em minimal} strategies for object manipulation, given the constraints of micro-robotic control strategies. In the spirit of minimality, we aim to tailor our low-level control policies to the natural dynamical behaviors of our system prior to the specification of increasingly abstract control policies. To this end, we have chosen to study ``bouncing robots,'' which have been shown to exhibit several behaviors, such as highly robust limit cycles, chaotic behavior, and large basins of attraction~\cite{alam2017minimalist,Nilles2018,spagnolie2017microorganism}. Once discovered, these behaviors can be chained together and leveraged towards solving robotic tasks such as coverage and localization without exceedingly complex control strategies~\cite{bayuelo2019computing}. The key insight in this paper is that by taking advantage of spontaneous limit cycles in the system dynamics, trajectories can be engineered that, purely as a result of the incidental collisions of the robot, manipulate objects in the robot's environment. Figure~\ref{fig:env_tri}(b) shows an example of such a trajectory constructed from iterative executions of the natural cyclic behavior of the bouncing robots. A detailed analysis of this dynamical system and its limit cycles are presented in Section~\ref{sec-dyn}.

\vspace{-0.2in}
\section{A Formal Comparison of Several Robot Designs}
\label{sec-designs}
\vspace{-0.1in}

In order to achieve the goal of manipulating an object in a long corridor, we will introduce several robot designs and then construct policies that each robot might use to accomplish the task. Particularly, these robots were designed to achieve the limit cycle behavior of bouncing robots, and to use this cyclic motion pattern to push the object in a specified direction.

\begin{figure}[t]
\centering
\includegraphics[width=.8\linewidth]{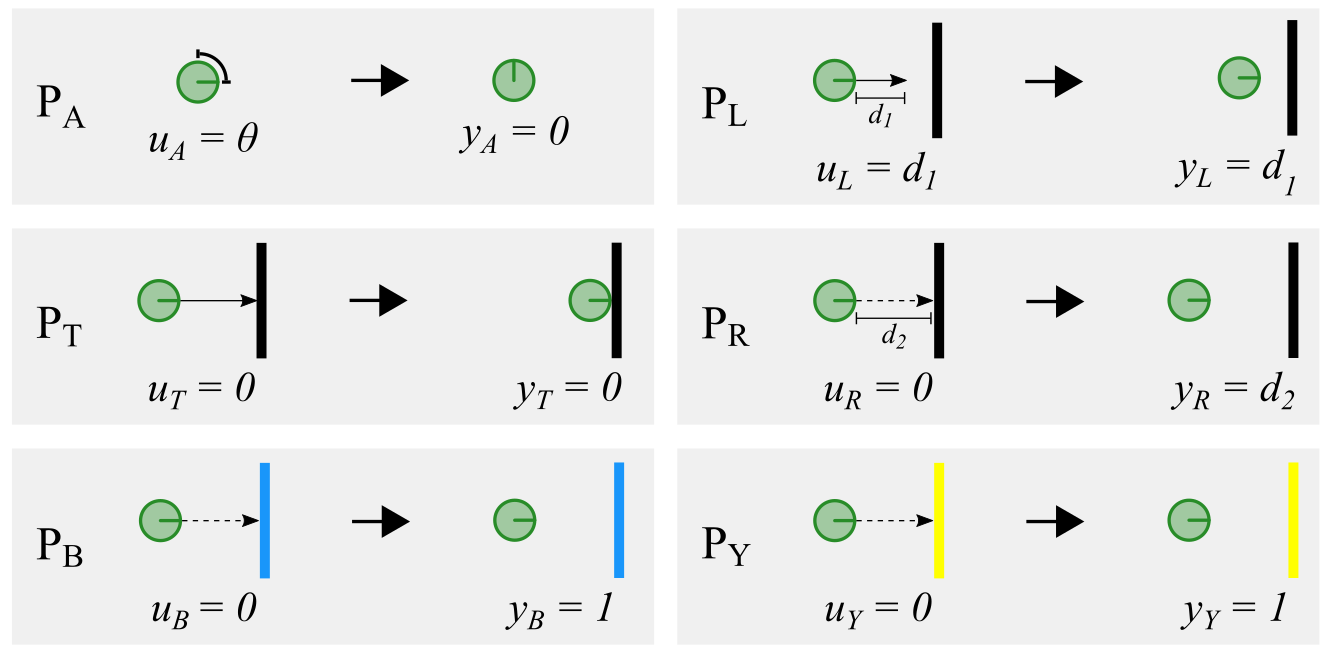}
\caption{\label{fig:primitives} Simple robotic behaviors called primitives. $P_A$ is a local rotation, $P_T$ is a forward translation to an obstacle, $P_L$ is a forward translation a set distance, $P_R$ is a range sensor, $P_B$ senses the color blue, and $P_Y$ senses the color yellow.}\vspace{-0.2in}
\end{figure}

Prior to describing the robot designs, we introduce the relevant primitives that will be used to construct the robots. Figure~\ref{fig:primitives} shows four robotic primitives taken directly from~\cite{okane2006} ($P_A$, $P_L$, $P_T$, and $P_R$) and two additional primitives defined for the proposed task ($P_B$ and $P_Y$). The primitive $P_A$ describes a rotation relative to the local reference frame given an angle $u_A$. $P_L$ corresponds to a forward translation over a chosen distance $u_L$, and $P_T$ carries out forward translation in the direction of the robot's heading until it reaches an obstacle. In addition to these actuation primitives, we define $P_R$ as a range sensor where $y_R$ is the distance to whatever is directly in front of the robot. For the 4-tuple specification of these primitives we refer the reader to~\cite{okane2006}. The chosen primitives were largely selected based on their feasibility of implementation at the micro-scale, as observed in many biological systems~\cite{spagnolie2017microorganism,kantsler2013ciliary}.

To differentiate different facets of the object being manipulated, we employ two primitives that are sensitive to the signature of each side of the target. $P_B$ is a blue sensor that measures $y_B = 1$ if the color blue is visible (in the geometric sense) given the robot's configuration $(x,y,\theta)$, and $P_Y$ is a yellow sensor outputting $y_Y = 1$ if the color yellow is visible to the robot. More formally, we define $P_B=(0,\{0,1\},f_B,h_B)$ and $P_Y=(0,\{0,1\},f_Y,h_Y)$, where $f_B$ and $f_Y$ are trivial functions always returning $0$, and the observation functions $h_B$ and $h_Y$ return $1$ when the appropriate signature is in front of the robot. We deliberately make the ``blue'' and ``yellow'' sensors abstract since they should be thought of as placeholders for any sensing capable of breaking the symmetry of the manipulation task, such as a chemical comparator, as discussed in Section~\ref{sec-motivation}.

These six simple primitives, shown in Fig.~\ref{fig:primitives} can be combined in different ways to produce robots of differing capabilities. Notably, here we develop modular subroutines and substrategies that allow us to develop hierarchical robot designs, enabling straightforward analysis of their capabilities. The corresponding task performance and design complexity trade-offs between our proposed robots and their constructed policies are explored in the following subsections.

\vspace{-0.15in}
\subsection{Robot 0: Omniscient and Omnipotent}
\vspace{-0.05in}
In many reported examples in micro-robot literature, the robots---as understood through the outlined framework---are \emph{not} minimal. Often, instead of grappling with the constraints of minimal on-board computation, designers make use of external sensors and computers to observe micro-robot states, calculate optimal actions, and actuate the micro-robots using external magnetic fields, sound waves, or other methods~\cite{Li2017,Xu2015,Xu2019}. Given the prevalence of such powerful robots in the literature, we introduce a ``perfect'' robot to demonstrate notation and compare to the minimal robots we present in the following sections.

We can specify the primitive for an all-capable robot as $P_O = (SE(2),SE(2)\times SE(2),f_O,h_O)$, where the action set $U_O$ is the set of all possible positions and orientations in the plane, and the observation set $Y_O$ is the set of all possible positions and orientations in the plane of both the robot and the object. The state transition function is $f_O(x,u)=(x+ u_{O_x}\Delta t_k ,y+u_{O_y}\Delta t_k ,\theta + u_{O_{\theta}}\Delta t_k )$ where $u_{O_x},u_{O_{y}},u_{O_{\theta}} \in \mathbb{R}$, and $\Delta t_k\in \mathbb{R}^+$ is the time step corresponding to the discrete amount of time passing between each stage $k$. The observation function outputs the current configurations of the robot and object. A robot with access to such a primitive (\textit{i.e.}, through external, non-minimal computing) would be able to simultaneously observe themselves and the object anywhere in the configuration space at all times and dexterously navigate to any location in the environment. Given that the configuration space of the robot is some bounded subset $X$ of $SE(2)$, the robot definition for this omniscient robot is $R_0 = (X,SE(2),SE(2)\times SE(2),f_O,h_O)$. Since all robots employed in the task of manipulating the cart in the long corridor share the same bounded configuration space  $X\subseteq SE(2)$, we also can define this robot using $R_0=\{P_O\}$. We use this alternative notation for robot definitions as it is less cumbersome and highlights differences in capabilities. While it is clear that such a robot should be able to solve the task through infinitely many policies, we provide an example of such a policy $\pi_0$ that completes the task in Algorithm~\ref{alg:r0} in the Appendix.

\vspace{-0.15in}
\subsection{Robot 1: Complex}
\vspace{-0.05in}
\label{sec-robot1}

\begin{figure}[t]
\centering
\includegraphics[width=0.95\linewidth]{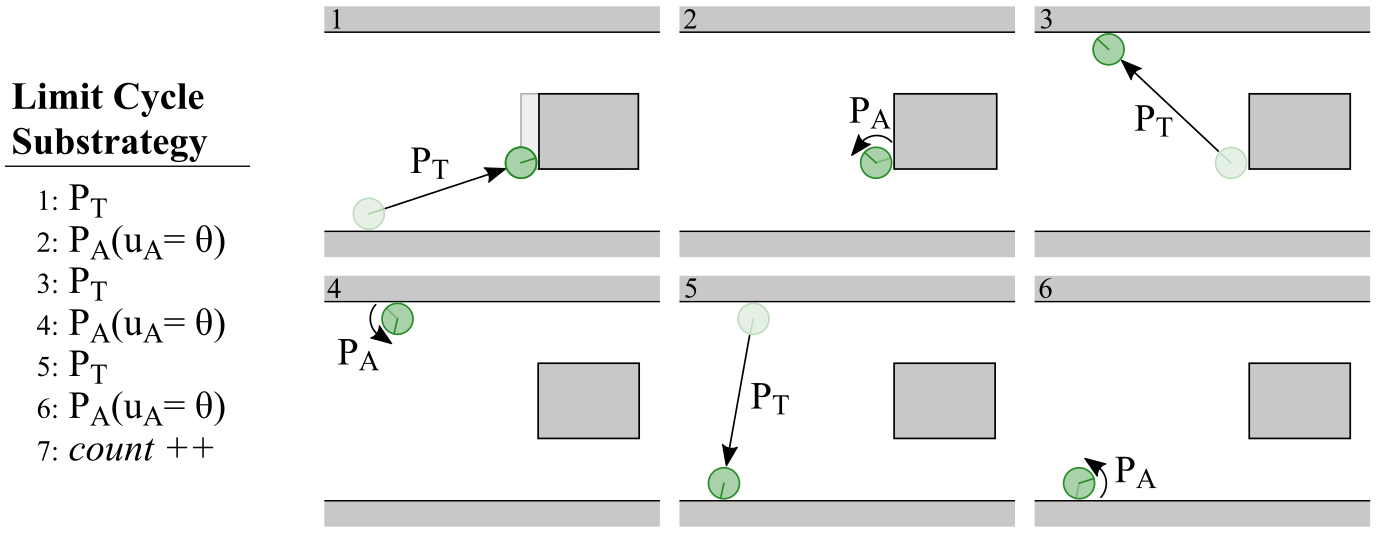}
\caption{\label{fig:LC} The \textit{Limit Cycle} strategy uses the primitive $P_T$, which moves forward until coming into contact with an object, and $P_A$, which rotates relative to the robot heading. Once a limit cycle is completed, the $count$ variable is incremented.}\vspace{-0.2in}
\end{figure}

\begin{figure}[tp!]
\centering
\includegraphics[width=0.95\linewidth]{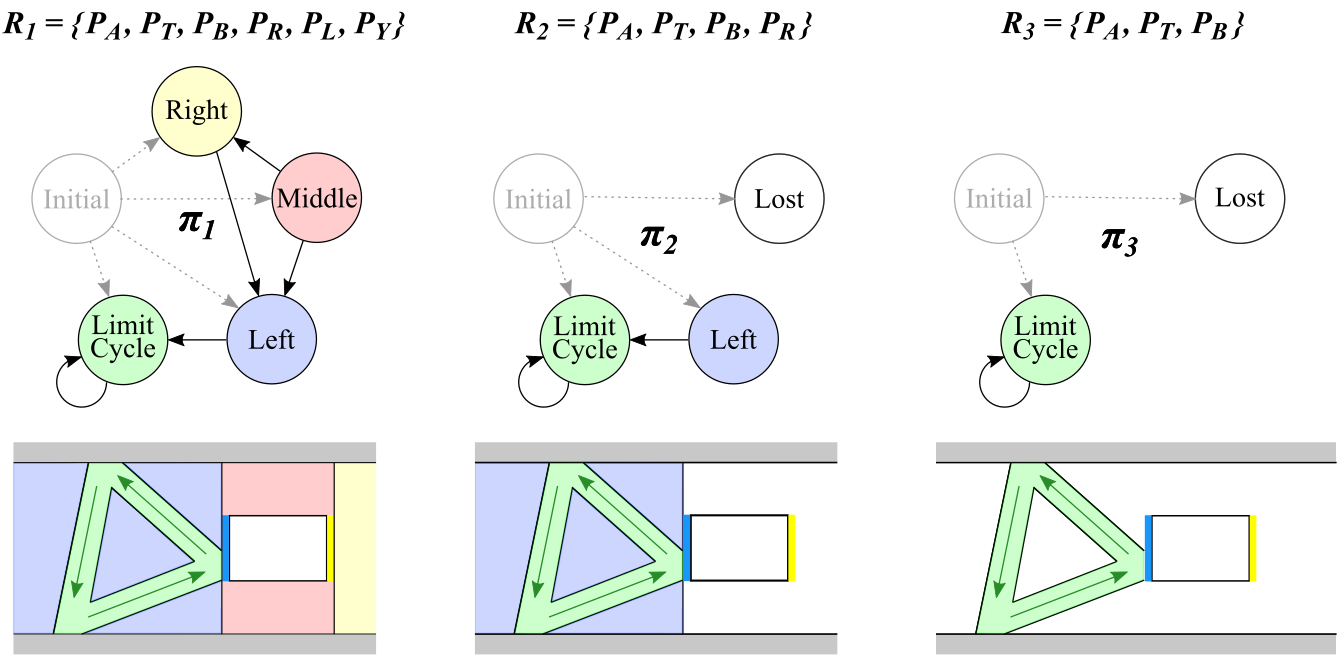}
\caption{\label{fig:strategies} (Left) A complex robot (composed of $6$ primitives) can successfully achieve its goal no matter its initial conditions. (Middle) A simple robot (composed of $4$ primitives) can only be successful if its initial conditions are on the left side of the object. (Right) A minimal robot (composed of $3$ primitives) can only be successful if its initial conditions are within the range of the limit cycle.}\vspace{-0.25in}
\end{figure}

The underlying design principle behind each of the following minimal robotic designs is modularity. We use a hierarchical control approach: at the highest level, we have a few spatio-temporal states (see Fig.~\ref{fig:strategies}). In each state, we use our available primitives to develop \textit{subroutines} corresponding to useful behaviors such as wall following, measuring distance to an object, and orienting a robot in the direction of the blue side of the cart. These subroutines are specified in the Appendix. Moreover, through these subroutines we construct \emph{substrategies} that transition the robots between states, such as moving from the right hand side of the object to the left hand side. We represent the complete robot policy $\pi_i$ as a combination of these substrategies in the form of a finite state machine (FSM).

The key substrategy that enables the success of our minimal robot designs is the \textit{Limit Cycle} substrategy (shown in Fig.~\ref{fig:LC}), which requires two primitives, $P_A$ and $P_T$, to perform. This substrategy is enabled by the fact that these bouncing robots converge to a limit cycle for a non-zero measure set of configurations (as shown in Section~\ref{sec-dyn}). All other substrategies employed in the design of our robots serve the purpose of positioning the robot into a configuration where it is capable of carrying out the limit cycle substrategy.

Hence, we define our robot as $R_1 = \{P_A, P_T, P_B, P_R, P_L, P_Y\}$, which makes use of all six of the primitives shown in Fig.~\ref{fig:primitives}. Its corresponding policy $\pi_1$, as represented by an FSM, is shown in Fig.~\ref{fig:strategies}. The details describing this policy are in Algorithm~\ref{alg:r1} in the Appendix. We highlight that through this policy the robot is capable of succeeding at the task from \emph{any} initial condition. The substrategy structure of the FSM---containing \textit{Initial}, \textit{Left}, \textit{Right}, \textit{Middle} and \textit{Limit Cycle} states---corresponds to different configuration domains that the robot may find itself in as represented by the shaded regions in Fig.~\ref{fig:strategies}. Due to the structure of the FSM and the task at hand, if the robot is in the \textit{Limit Cycle} state it will eventually succeed at the task. Finally, although we allocate memory for the robot to track its success through a variable \textit{count} (as seen in the substrategy in Fig.~\ref{fig:LC}) the robot does not require memory to perform this substrategy and we have only included it for facilitating analysis. 

\vspace{-0.2in}
\subsection{Robot 2: Simple}
\vspace{-0.05in}
\label{sec-robot2}
Robot 2, defined as $R_2 = \{P_A, P_T, P_B, P_R\}$, is comprised of a subset of the primitives from $R_1$. As a result, it is not capable of executing all of the same motion plans as $R_1$. As Fig.~\ref{fig:strategies} shows, $R_2$ can enter its \textit{Limit Cycle} substrategy if it starts on the left side of the object, but otherwise it will get lost. The \textit{Initial} state uses sensor feedback to transition to the substrategy the robot should use next. The \textit{Lost} state is distinct from the \textit{Initial} state---once a robot is lost, it can never recover (in this case, the robot will move until it hits a wall and then stay there for all time). More details on policy $\pi_2$ and the specific substrategies that $R_2$ uses can be found in Algorithm~\ref{alg:r2} in the Appendix.

\vspace{-0.15in}
\subsection{Robot 3: Minimal}
\vspace{-0.05in}
\label{sec-robot3}

Robot 3, $R_3 = \{P_A, P_T, P_B\}$, contains three robotic primitives, which are a subset of the primitives of $R_1$ and $R_2$. Under policy $\pi_3$ (detailed in Algorithm~\ref{alg:r3} in the Appendix), this robot can only be successful at the task if it initializes in the \textit{Limit Cycle} state, facing the correct direction. Otherwise, the robot will never enter the limit cycle. Such a simple robot design could be useful in a scenario when there are very many ``disposable" robots deployed in the system. Even if only a small fraction of these many simple robots start out with perfect initial conditions, the goal would still be achieved. Despite the apparent simplicity of such a robot, we note that $R_3$ (along with all other introduced designs) is capable of determining whether or not it is succeeding at the task or whether it is lost irreversibly. Such capabilities are not by any means trivial, but are included in the robot designs for the purposes of analysis and comparison.

\vspace{-0.2in}
\subsection{Comparing Robots}
\vspace{-0.05in}
\label{sec-comparison}
We will compare the four robots introduced in this section, $R_0,\ R_1,\ R_2$, and $R_3$. In order to achieve this we must first specify the task and derived I-space in which we can compare the designs. The chosen derived I-space is $\mathcal{I}_{der} = \mathbb{Z}^+\cup\{0\}$. Specifically, it consists of counts of the \textit{Limit Cycle} state (the $count$ variable is shown in Fig.~\ref{fig:LC} and in the algorithms in the Appendix). If we assume that after each collision the robot pushes the object a distance $\epsilon$, task success is equivalently tracked in memory by $count$ up to a scalar.

We express the goal for the task of manipulating the cart in a long corridor as $\mathcal{I}_G\subseteq\mathcal{I}_{der}$, where $\mathcal{I}_G$ is an open subset of the nonnegative integers. In this set up, as illustrated in Fig.~\ref{fig:env_tri}(a), the robot must push the object some $N$ times, corresponding to a net distance traveled, to succeed. More formally, we express our information preference relation through the indicator $1_G(\eta)$ corresponding to whether a derived information history is within the goal region $\mathcal{I}_G$, thereby inducing a partial ordering over information states. Hence, the likelihood of success of any of the proposed robot designs (excluding $R_0$) is solely determined by their initialization, and the region of attraction of the limit cycle behavior for the bouncing robots, which will be explored in more detail.

\vspace{-0.2in}
\subsubsection{Comparing $R_1$, $R_2$, and $R_3$}
The comparison of robots $R_1$, $R_2$, and $R_3$ through the lens of robot dominance is straightforward given our modular robot designs. Since $R_1$ and $R_2$ are comprised of a superset of the primitives of $R_3$, they are strictly as capable or more capable than $R_3$, as per Lemma 1(a). Therefore, we state that $R_1$ and $R_2$ \emph{dominate} $R_3$, denoted by $R_3 \unlhd R_1$, and $R_3 \unlhd R_2$. Likewise, using the same lemma, we can see that $R_2 \unlhd R_1$. This is to say that for the task of manipulating the cart along the long corridor $R_1$ should outperform $R_2$ and $R_3$, and that $R_2$ should outperform $R_3$.

While the policies for each robot design are nontrivial, Fig.~\ref{fig:strategies} offers intuition for the presented dominance hierarchies. Effectively, if either $R_2$ or $R_3$ are initialized into their \textit{Lost} state they are incapable of executing the task for all time. Hence, it is the configuration space volume corresponding to the \textit{Lost} state that determines the robot dominance hierarchy. 

Let $\eta^{(1)}\in\mathcal{I}_{hist}^{(1)}, \eta^{(2)}\in\mathcal{I}_{hist}^{(2)}, \eta^{(3)}\in\mathcal{I}_{hist}^{(3)}$, and define I-maps that return the variable $count$ stored in memory for each robot. The information preference relation then only discriminates whether the information histories correspond to a trajectory reaching $\mathcal{I}_G\subseteq\mathcal{I}_{der}$---in other words, whether a robot achieves the required $N$ nudges to the object in the corridor. Note that since there is no time constraints to the task, this number is arbitrary and only relevant for tuning to the length-scales of the problem. Thus, the dominance relations outlined above follow from the fact that for non-zero volumes of the configuration space there exists no integer $l$ for which $\kappa_1(\eta^{(1)})\preceq\kappa_2(F^l(\eta^{(2)},\pi_2,x))$. On the other hand for all $x\in X$, $\kappa_2(\eta^{(2)})\preceq\kappa_1(F^l(\eta^{(1)},\pi_1,x))$. Through this same procedure we can deduce the rest of the hierarchies presented in this section.

\vspace{-0.2in}
\subsubsection{Comparing $R_0$ and $R_1$}
To compare $R_0$ and $R_1$ we may continue with a similar reachability analysis as in the previous subsection. We note that from any $x\in X$, $R_1$ and $R_0$ are capable of reaching the object and nudging it. This means that given that the domain $X$ is bounded and information history states $\eta^{(0)}\in\mathcal{I}_{hist}^{(0)},\eta^{(1)}\in\mathcal{I}_{hist}^{(1)}$ corresponding to each robot, there always exists a finite integer $l$ such that $\kappa_0(\eta^{(0)})\preceq\kappa_1(F^l(\eta^{(1)},\pi_1,x))$, and $\kappa_1(\eta^{(1)})\preceq\kappa_0(F^l(\eta^{(0)},\pi_0,x))$. So we have that $R_1 \unlhd R_0$ and $R_0 \unlhd R_1$, meaning that $R_1 \equiv R_0$. Thus, $R_0$ and $R_1$ are equivalently capable of performing the considered task. 

It is important to note that despite the intuition that $R_0$ is more ``powerful'' than $R_1$ in some sense, for the purposes of the proposed task that extra power is redundant. However, there are many tasks where this is would not be the case (\textit{e.g.}, moving the robot to a specific point in the plane).

\vspace{-0.2in}
\subsubsection{Comparing $R_0$, $R_2$ and $R_3$}
Lastly, while the relationship between $R_0$ and the other robot designs is intuitive, we must introduce an additional lemma.

\noindent\textbf{Lemma 2.} (Transitive property) Given three robots $R_0,R_1,R_2$, if $R_2\unlhd R_1$ and $R_1 \equiv R_0$, then $R_2\unlhd R_0$.

\noindent\textit{Proof.} The proof of the transitive property of robot dominance comes from the definition of equivalence. $R_1 \equiv R_0$ means that the following statements are simultaneously true: $R_1 \unlhd R_0$ and $R_0 \unlhd R_1$. Thus, this means that $R_2 \unlhd R_1 \unlhd R_0$, which implies that $R_2 \unlhd R_0$, concluding the proof.

Using this additional lemma, we see that $R_2 \unlhd R_0$, and $R_3 \unlhd R_0$, as expected. Hence, we have demonstrated that minimal robots may be capable of executing complex strategies despite the constraints imposed by the micro-scale domain. Minimality in micromanipulation is in fact possible when robot designs take advantage of naturally occurring dynamic structures, such as limit cycles. In the following sections we discuss the necessary conditions for establishing such cycles, as well as the robustness properties of limit cycle behavior, which are important for extending this work to less idealized and deterministic settings.  

\vspace{-0.1in}
\section{Feasibility and Dynamics of Cyclic Motion Strategies}
\vspace{-0.0in}
\label{sec-dyn}

In this section we will derive and analyze limit cycle motion strategies that can be used to manipulate objects through incidental collisions. The goal is to engineer robust patterns in the robot's trajectory that are useful for this task. When looking to move an object down the corridor, three boundaries are present: the two walls of the corridor, and the object itself. An ideal motion strategy would use collisions with the two walls to direct the robot to collide with the object in a repeatable pattern. Is it possible to do so with a single instruction to the robot that it repeats indefinitely, every time it encounters a boundary? This single-instruction strategy would lend itself well to the design of a micro-robot, so that the robot robustly performs the correct boundary interaction each time. 

\begin{figure}[pt] \center
\includegraphics[width=.95\linewidth]{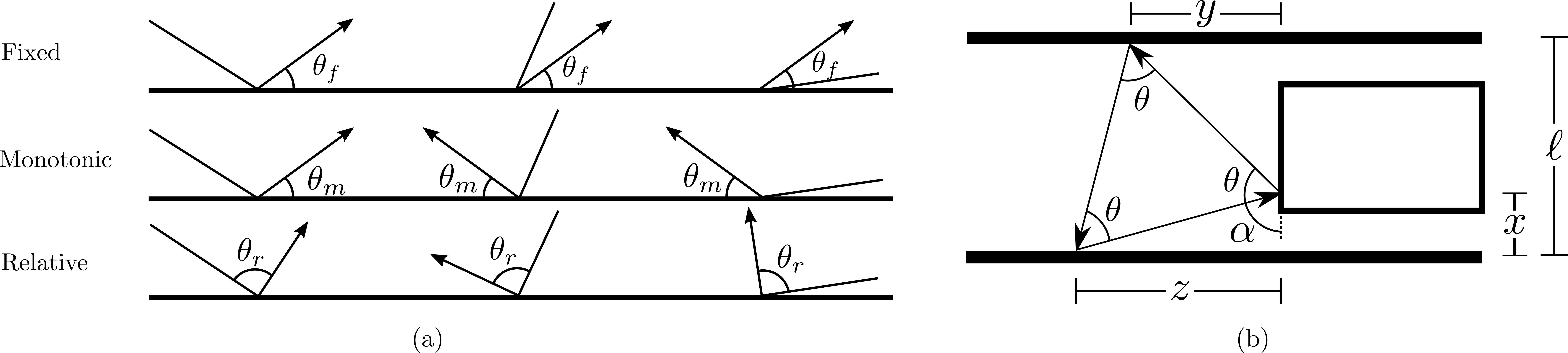}
\caption{\label{fig:bounce_types} (a) The three types of bounce rules considered. The top row depicts {\em fixed} bounce rules, where the robot leaves the boundary at a fixed angle regardless of the incoming trajectory. The second row shows {\em fixed monotonic} bounce rules preserving the horizontal direction of motion but keep the absolute angle between the boundary and the outgoing trajectory fixed. The third row shows {\em relative} bounce rules rotating the robot through an angle relative to its previous heading. (b) The geometric setup for analyzing dynamics of triangular trajectories formed by repeated single-instruction relative bounce rules.}\vspace{-0.2in}
\end{figure}

We consider three types of boundary actions, as seen in Fig.~\ref{fig:bounce_types}(a). The first and second types (fixed bounce rules) could be implemented through alignment with the boundary (mechanical or otherwise) such that forward propulsion occurs at the correct heading. This measurement and reorientation can be done compliantly, and does not necessarily require traditional onboard measurement and computation. See, for example, similar movement profiles of microorganisms resulting from body morphology and ciliary contact interactions~\cite{kantsler2013ciliary,spagnolie2017microorganism}. The third type of boundary interaction (relative bounce rule) requires a rotation relative to the robot's prior heading, implying the need for rotational odometry or a fixed motion pattern triggered upon collision.

Here, we analyze the relative power of these actions for the task of pushing an object down a hallway, without considering the broader context of initial conditions or localization, which were considered in Section~\ref{sec-designs}. Particularly, we consider the system of the robot, hallway and object as a purely dynamical system, to establish the feasibility of trajectories resulting from minimal policies.

For the rest of this section, suppose that the bouncing robot navigates in a corridor with parallel walls and a rectangular object as described in Section~\ref{sec-motivation}.

\vspace{-0.05in}
\begin{proposition}
For all bouncing robot strategies consisting of a single repeated fixed bounce rule, there does not exist a strategy that would result in a triangular trajectory that makes contact with the rectangular object.
\label{prop-bounce}
\end{proposition}

\vspace{-0.1in}
\begin{proof}
A fixed bounce rule in this environment will result in the robot bouncing back and forth between the two parallel walls forever after at most one contact with the object. A monotonic fixed bounce rule would result in the robot bouncing down the corridor away from the object after at most one contact.
\end{proof}

\vspace{-0.1in}
\begin{remark}
The feasibility of fixed (monotonic) bounce rule strategies in environments without parallel walls is unknown and possibly of interest.
\end{remark}

\vspace{-0.1in}
\begin{proposition}
There exist an infinite number of strategies consisting of two fixed bounce rules that each result in a triangular trajectory that makes contact with the rectangular object. The robot must also be able to distinguish the object from a static boundary, or must know its initial conditions.
\label{prop-fixed}
\end{proposition}

\vspace{-0.1in}
\begin{proof}
Geometrically, an infinite number of triangles exist that can be executed by a robot with the choice between two different fixed bounce rules. It is necessary that the robot be able to determine when it has encountered the first corridor wall during its cycle, in order to switch bounce rules to avoid the situation described in the proof of Proposition~\ref{prop-bounce}. One sufficient condition is that the robot knows the type of boundary at first contact, and has one bit of memory to track when it encounters the first corridor wall and should switch actions. Equivalently, a strategy could use a sensor distinguishing the object from a static boundary at collision time, along with one bit of memory.
\end{proof}

\vspace{-0.1in}
\begin{proposition}
There exists a strategy consisting of a single repeated relative bounce rule that results in a triangular trajectory that makes contact with the rectangular object. Moreover, this strategy is robust to small perturbations in the rotation $\theta$ and the initial angle $\alpha$.
\label{prop-relative}
\end{proposition} 

\begin{figure}[pt]\center
\includegraphics[width=0.9\linewidth]{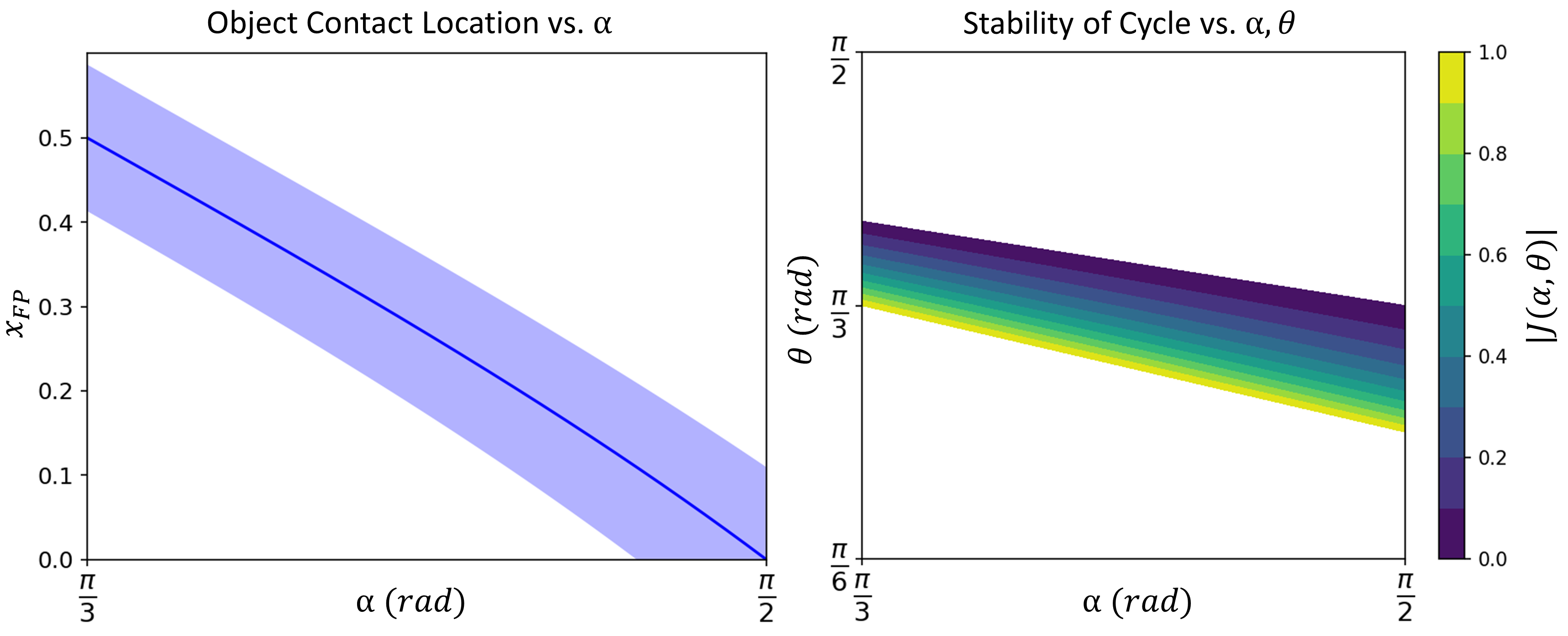}
\caption{(Left) The location of the impact on the object is nearly a linear function of $\alpha$ for $\theta=\frac{\pi}{3}\pm0.1$. Only the counterclockwise cycle is shown; the clockwise cycle follows from symmetry.  (Right) The Jacobian of this system indicates robustness to small perturbations in $\theta$ (white regions represent instability or infeasibility).}
\label{fig:robustness}
\vspace{-0.2in}
\end{figure}

\vspace{-0.1in}
\begin{proof}
See Fig.~\ref{fig:bounce_types}(b) for the geometric setup. Here we will provide exact expressions for the quantities $x,y$ and $z$ as a function of the initial conditions, position, and orientation of the robot on its first collision with the object.

First assume $x_k$ is given, as the point of impact of the robot with the object at stage $k$. Let $\alpha$ be the angle indicated in Fig.~\ref{fig:bounce_types}(b), the angle between the incoming trajectory at $x_k$ and the object face. Then, using simple trigonometry, $y = (\ell - x_k) \tan(\pi - \theta - \alpha)$ where $\theta$ is the interior angle of the robot's rotation. To create an equiangular triangle, $\theta = \frac{\pi}{3}$, but we will leave $\theta$ symbolic for now to enable sensitivity analysis. To compute $z$, we consider the horizontal offset due to the transition from the top to the bottom of the hallway. This leads to $z = y + \ell\cot(\frac{3\pi}{2} - 2\theta - \alpha)$. Finally, we can compute the coordinate of where the robot will return to the object, $x_{k+3} = z \tan(\frac{3\pi}{2} - \alpha - 3\theta)$. Solving for the fixed point of this dynamical system, $x_{FP} = x_{k} = x_{k+3}$ gives
\vspace{-0.1in}
$$
x_{FP} = \ell \left(
\frac{\tan(\alpha+\theta) - \tan(\alpha+2\theta)}{\tan(\alpha+\theta)-\tan(\alpha+3\theta)} \right).
$$ \vspace{-0.15in}

\noindent Note that this $x_{FP}$ expression is not valid for all values of $(\alpha,\theta)$, but only for values leading to inscribed triangles in the environment such that $x_{FP}>0$. 

For $\theta = \frac{\pi}{3}$, Fig.~\ref{fig:robustness} (Left) shows the location of the fixed point as a function of the angle $\alpha$, indicating that an infinite number of stable cycles exist for this motion strategy, and every point of the object in the bottom half of the corridor is contactable with counterclockwise cycles. The top half of the object can be reached using clockwise cycles.

\end{proof}
\vspace{-0.1in}

Since $x_{k+3} = F(x_k,\alpha,\theta)$ is linear in $x_k$, the Jacobian is $J = -\tan(\alpha+\theta)\tan(\alpha+3\theta-\frac{\pi}{2})$. Figure \ref{fig:robustness} (Right) shows the value of the Jacobian as a function of $\alpha$ and $\theta$. All the shaded regions have an absolute value less than one, indicating robustness to small perturbations of $\alpha$ and $\theta$ over a large domain.

\vspace{-0.1in}
\begin{remark}
We note that Propositions \ref{prop-fixed} and \ref{prop-relative} are equivalent in terms of memory requirements, both requiring the robot to ``remember'' either its previous heading or its previous bounce rule at each stage of the strategy. However, the fixed bounce strategy implies that the robot has more knowledge of its surroundings than the relative bounce strategy, as it must measure the plane of the boundary it encounters and orient itself appropriately. The relative bounce strategy requires only information within the robot's own reference frame.
\end{remark}

\vspace{-0.25in}
\section{Conclusion}
\vspace{-0.1in}

In this paper, we have designed robust motion strategies for minimal robots that have great promise for micromanipulation. We have also analyzed the information requirements for task success, compared the capabilities of four different robot designs, and found that minimal robot designs may still be capable of micromanipulation without the need for external computation. While our example of a rectangular obstacle in a corridor is simple, we think of this as \emph{robust directed transport}, a key building block for future work.

\vspace{-0.15in}
\subsection{Future Directions}
\vspace{-0.05in}
The setting of micro-robotics provides motivation for the approach we have laid out in this work. At the micro-scale, coarse high-level controllers that can be applied to a collection of many micro-robots are easier to implement than fine-grained individual controllers. This requires formal reasoning about all possible trajectories, in order to funnel the system into states that allow for task completion, as was illustrated in this work. In order to be more applicable in the micro-robotics domain, it will be important to extend the approach to multiple agents, as well as scenarios subject to noise. While our strategies passively provide some noise tolerance by virtue of the limit cycle region of attraction, there is much work to be done on more concrete applications to characterize and account for sensing and actuation noise.

Outside of our particular model and application, this work has implications for the future of robot behavior and design. Often, derived I-states are designed to infer information such as the set of possible current states of the robot, or the set of possible states that the robot could have previously occupied. In this work, we focus on derived I-spaces that encode information about what {\em will happen} to the robot under a given strategy. With these forward-predictive derived I-spaces, we encode a high density of task-relevant information into a few-state symbolic abstraction. By making use of such abstractions, minimal agents may be endowed with a passively predictive capacity leading to greater task-capability.

More broadly, this work provides an exciting glimpse toward more automated analysis, through a combination of system identification techniques, hybrid systems theory, and I-space analysis. Coarse-grained sensors provide an avenue for discretization useful for hierarchical control; such an approach is increasingly needed as our robotic systems become more data-driven. Such a unified approach may be able to simultaneously identify coarse-grained system dynamics, predict their task-capabilities, and design fine-tuned control strategies. 

\vspace{-1em}

\bibliographystyle{plain}
\bibliography{main}

\begin{thebibliography}{10}

\bibitem{alam2017minimalist}
T.~Alam, L.~Bobadilla, and D.~A. Shell.
\newblock Minimalist robot navigation and coverage using a dynamical system
  approach.
\newblock In {\em IEEE Int. Conf. Rob. Comp.}, 2017.

\bibitem{Alam2018}
T.~{Alam}, L.~{Bobadilla}, and D.~A. {Shell}.
\newblock Space-efficient filters for mobile robot localization from discrete
  limit cycles.
\newblock {\em IEEE Rob. Auto. Lett.}, 3(1):257--264, 2018.

\bibitem{bayuelo2019computing}
A.~Bayuelo, T.~Alam, L.~Bobadilla, L.~F. Ni{\~n}o, and R.~N. Smith.
\newblock Computing feedback plans from dynamical system composition.
\newblock In {\em IEEE Int. Conf. Auto. Sci. Eng.}, pages 1175--1180, 2019.

\bibitem{Censi2016}
A.~Censi.
\newblock A mathematical theory of co-design.
\newblock Technical report, September 2016.
\newblock Conditionally accepted to IEEE Trans. Rob.

\bibitem{Chambers1931}
R.~Chambers, H.~B. Fell, and W.~B. Hardy.
\newblock Micro-operations on cells in tissue cultures.
\newblock {\em Proc. Royal Soc. of London.}, 109(763):380--403, 1931.

\bibitem{Ding2013}
J.~Ding, V.~R. Challa, M.~G. Prasad, and F.~T. Fisher.
\newblock {\em Vibration Energy Harvesting and Its Application for Nano- and
  Microrobotics}.
\newblock Springer, New York, NY, 2013.

\bibitem{donald1997}
B.~R. Donald, J.~Jennings, and D.~Rus.
\newblock Information invariants for distributed manipulation.
\newblock {\em Int. J. Rob. Res.}, 16(5):673--702, 1997.

\bibitem{Douglas2012}
S.~M. Douglas, I.~Bachelet, and G.~M. Church.
\newblock A logic-gated nanorobot for targeted transport of molecular payloads.
\newblock {\em Science}, 335(6070):831--834, 2012.

\bibitem{Ilton2018}
M.~Ilton, M.~S. Bhamla, X.~Ma, S.~M. Cox, L.~L. Fitchett, Y.~Kim, J.~Koh,
  D.~Krishnamurthy, C.~Kuo, F.~Z. Temel, A.~J. Crosby, M.~Prakash, G.~P.
  Sutton, R.~J. Wood, E.~Azizi, S.~Bergbreiter, and S.~N. Patek.
\newblock The principles of cascading power limits in small, fast biological
  and engineered systems.
\newblock {\em Science}, 360(6387), 2018.

\bibitem{kantsler2013ciliary}
Vasily Kantsler, J{\"o}rn Dunkel, Marco Polin, and Raymond~E Goldstein.
\newblock Ciliary contact interactions dominate surface scattering of swimming
  eukaryotes.
\newblock {\em Proc. Natl. Acad. Sci.}, 110(4):1187--1192, 2013.

\bibitem{kim2015}
J.~Kim and D.~A. Shell.
\newblock A new model for self-organized robotic clustering: Understanding
  boundary induced densities and cluster compactness.
\newblock In {\em IEEE Int. Conf. Rob. Auto. (ICRA)}, pages 5858--5863, May
  2015.

\bibitem{LaValle2006}
S.~M. LaValle.
\newblock {\em Planning Algorithms}.
\newblock Cambridge University Press, USA, 2006.

\bibitem{Li2017}
J.~Li, B.~Esteban-Fern{\'a}ndez~de {\'A}vila, W.~Gao, L.~Zhang, and J.~Wang.
\newblock Micro/nanorobots for biomedicine: Delivery, surgery, sensing, and
  detoxification.
\newblock {\em Sci. Rob.}, 2(4), 2017.

\bibitem{Nilles2018}
A.~Q. Nilles, Y.~Ren, I.~Becerra, and S.~M. LaValle.
\newblock A visibility-based approach to computing nondeterministic bouncing
  strategies.
\newblock {\em Int. Work. Alg. Found. Rob.}, Dec. 2018.

\bibitem{okane2006}
J.~M. O'Kane and S.~M. LaValle.
\newblock Comparing the power of robots.
\newblock {\em Int. J. Rob. Res.}, 27(1):5--23, 2008.

\bibitem{Oudenhoven2011}
J.~F.~M. Oudenhoven, L.~Baggetto, and P.~H.~L. Notten.
\newblock All-solid-state lithium-ion microbatteries: A review of various
  three-dimensional concepts.
\newblock {\em Adv. Energy Mat.}, 1(1):10--33, 2011.

\bibitem{Pervan2018}
A.~Pervan and T.~D. Murphey.
\newblock Low complexity control policy synthesis for embodied computation in
  synthetic cells.
\newblock {\em Int. Work. Alg. Found. Rob.}, Dec. 2018.

\bibitem{Saberifar2018}
F.~Z. Saberifar, J.~M. O'Kane, and D.~A. Shell.
\newblock The hardness of minimizing design cost subject to planning problems.
\newblock {\em Int. Work. Alg. Found. Rob.}, Dec. 2018.

\bibitem{Savoie2019}
W.~Savoie, T.~A. Berrueta, Z.~Jackson, A.~Pervan, R.~Warkentin, S.~Li, T.~D.
  Murphey, K.~Wiesenfeld, and D.~I. Goldman.
\newblock A robot made of robots: Emergent transport and control of a smarticle
  ensemble.
\newblock {\em Sci. Rob.}, 4(34), 2019.

\bibitem{Sitti2015}
M.~Sitti, H.~Ceylan, W.~Hu, J.~Giltinan, M.~Turan, S.~Yim, and E.~Diller.
\newblock Biomedical applications of untethered mobile milli/microrobots.
\newblock {\em Proc. of IEEE}, 103(2):205--224, Feb. 2015.

\bibitem{Soto2018}
F.~Soto and R.~Chrostowski.
\newblock Frontiers of medical micro/nanorobotics: in vivo applications and
  commercialization perspectives toward clinical uses.
\newblock {\em Front. in Bioeng. and Biotech.}, 6:170--170, Nov. 2018.

\bibitem{spagnolie2017microorganism}
S.~E. Spagnolie, C.~Wahl, J.~Lukasik, and J.~Thiffeault.
\newblock Microorganism billiards.
\newblock {\em Phys. D}, 341:33--44, 2016.

\bibitem{Xu2015}
T.~Xu, J.~Yu, X.~Yan, H.~Choi, and L.~Zhang.
\newblock Magnetic actuation based motion control for microrobots: An overview.
\newblock {\em Micromachines}, 6(9):1346--1364, Sept. 2015.

\bibitem{Xu2019}
T.~Xu, J.~Zhang, M.~Salehizadeh, O.~Onaizah, and E.~Diller.
\newblock Millimeter-scale flexible robots with programmable three-dimensional
  magnetization and motions.
\newblock {\em Sci. Rob.}, 4(29), 2019.

\bibitem{Yang2018}
G.~Z. Yang, J.~Bellingham, P.~E. Dupont, P.~Fischer, L.~Floridi, R.~Full,
  N.~Jacobstein, V.~Kumar, M.~McNutt, R.~Merrifield, B.~J. Nelson,
  B.~Scassellati, M.~Taddeo, R.~Taylor, M.~Veloso, Z.~L. Wang, and R.~Wood.
\newblock The grand challenges of science robotics.
\newblock {\em Sci. Rob.}, 3(14), 2018.

\end{thebibliography}

\newpage
\section*{Appendix}

\subsection*{Subroutines}
We define how the subroutines introduced in Section \ref{sec-designs} can be achieved. Subroutines for wall following (in a random direction) for $u_L$ steps, observing the object $y_R,y_B,y_Y$, and orienting in the direction of the blue side of the object are shown below, in Algorithm \ref{alg:subroutines}.

\begin{algorithm*}
\caption{Subroutines} \label{alg:subroutines}
\begin{algorithmic}

\STATE \textbf{wall\_follow} (input $u_L$)
\WHILE{$P_R(y_R \neq \infty)$ \hspace*{\fill}(while the robot is not parallel to the wall)}
    \STATE $P_A(u_A=\phi)$ \hspace*{\fill}(rotate a small angle $\phi$)
    \ENDWHILE
\STATE $P_L(u_L)$ \hspace*{\fill}(step forward a distance $u_L$)
\STATE \textcolor{white}{.}
\STATE \textcolor{white}{.}

\STATE \textbf{observe\_object} (output $y_R$, $y_B$, $y_Y$)
\STATE $inc = 0$
\WHILE{$P_B(y_B=0)$ and $P_Y(y_Y=0)$ and $inc \leq 360^\circ$ \hspace*{\fill}(while neither yellow nor blue}
\STATE \textcolor{white}{.}  \hspace*{\fill} are detected, and the robot has
\STATE \textcolor{white}{.}  \hspace*{\fill} not completed a full rotation)
    \STATE $P_A(u_A=\phi)$ \hspace*{\fill}(rotate a small angle $\phi$)
    \STATE $inc ++$
    \ENDWHILE
\IF{$y_B=1$ or $y_Y=1$ \hspace*{\fill}(if blue or yellow are detected)}
    \STATE $y_R = P_R$ \hspace*{\fill}(measure the distance to the color)
\ELSE
    \STATE $y_R=\infty$ \hspace*{\fill}(otherwise, return $\infty$ to encode `no color')
    \ENDIF

\STATE \textcolor{white}{.}
\STATE \textcolor{white}{.}

\STATE \textbf{aim\_toward\_blue} ()
\WHILE{$P_B(y_B = 0)$ \hspace*{\fill}(while blue is not detected)}
    \STATE $P_A(u_A=\phi)$ \hspace*{\fill}(rotate a small angle $\phi$)
    \ENDWHILE

\end{algorithmic}
\end{algorithm*}

For wall following, the robot continuously rotates a small angle (using the $P_A$ primitive) until it detects that there is nothing in front of it (when the range detecting primitive $P_R$ reads $\infty$), and is therefore facing a direction parallel to the wall. Then the robot uses primitive $P_L$ to move forward $u_L$ steps.

For observing the object, the robot continuously rotates until it has either detected the blue side of the object, detected the yellow side of the object, or completed a full rotation. If it has detected a color, it records the distance to the object and the color detected (if a color was detected). If the robot completes a full rotation without detecting either color, it returns a range of $\infty$, to encode that no color was found. It is possible to call a subset of measurements from this subroutine, for example in Algorithm~\ref{alg:r1} the \emph{Middle} substrategy only queries the observe$\_$object() subroutine for $y_B$ and $y_Y$.

For aiming toward blue, the robot will rotate in place (using $P_A$) until it detects the color blue (using $P_B$).

\subsection*{Robot 0 Policy}

Robot $0$ uses the primitive $P_O$, and requires knowledge of the width of the channel, $\ell$, and the length of the object, $2s$. First, in its initial state, it observes its own position $(x_r,y_r)$ and the position of the object $(x_o,y_o)$. If the robot, $x_r$, is to the left of object's left edge, $(x_o-s)$ (where $x_o$ is the center of mass of the object and $s$ is half the length of the object), then the robot should execute substrategy \emph{Left}.

If robot is to the right of the object's edge, it will go straight up, $u_{O_y}=\ell/\Delta t_k$, to either the top wall of the channel, or the underside of the object (if the robot happened to start below the object), and then move left, $u_{O_x}=((x_o-s)-x_r)/\Delta t_k$, until it has passed the object. Then the robot transitions to the \emph{Left} substrategy.

In the \emph{Left} substrategy, the robot translates to the left side of the object, and then pushes it a set distance $\epsilon$ to the right. It increases the $count$ variable with each subsequent push.

\begin{algorithm}
\caption{Robot 0 Policy: $\pi_0$} \label{alg:r0}
\begin{algorithmic}
\STATE \textbf{Requires} 
\STATE Primitive: $P_O$
\STATE Parameters: $s$ is half of the length of the object, $\ell$ is the width of the channel
\STATE \textcolor{white}{.}

\STATE \textbf{Initial}
\STATE $count = 0$
\STATE $x_r,y_r,x_o,y_o = P_O()$ \hspace*{\fill}(read the positions of the robot and the object)
\IF{$x_r<(x_o-s)$ \hspace*{\fill}(if the robot is to the left of the object's left edge)}
    \STATE Switch to \textbf{Left}
\ELSE{}
    \STATE $P_O(u_{O_x}=0,u_{O_y}=\ell/\Delta t_k)$ \hspace*{\fill}(move up, until the object or wall)
    \STATE $P_O(u_{O_x}=((x_o-s)-x_r)/\Delta t_k,u_{O_y}=0)$ \hspace*{\fill}(move to the left of the object)
    \STATE Switch to \textbf{Left}
\ENDIF

\STATE \textcolor{white}{.}
\STATE \textcolor{white}{.}

\STATE \textbf{Left}
\STATE $P_O(u_{O_x}=((x_o-s)-x_r)/\Delta t_k,u_{O_y}=(y_o-y_r)/\Delta t_k)$ \hspace*{\fill}(go to the center of the 
\STATE \textcolor{white}{.} \hspace*{\fill}left side of the object)
\STATE $P_O(u_{O_x}=\epsilon/\Delta t_k,u_{O_y}=0)$ \hspace*{\fill}(push the object a distance of $\epsilon$ to the right)
\STATE $count ++$

\end{algorithmic}
\end{algorithm}

\newpage
\subsection*{Robot 1 Policy}

Robot $1$ uses the primitives $P_A$, $P_T$, $P_B$, $P_R$, $P_L$, and $P_Y$, and requires knowledge of the set of distances $w$ from the object that allow the robot to fall into the limit cycle. First, in its initial state, $R_1$ measures the distance between it and the object, and the color directly in front of it (if any). It uses this knowledge to switch to a substrategy: either \emph{Limit Cycle}, \emph{Left}, \emph{Right}, or \emph{Middle}.

In the \emph{Limit Cycle} substrategy, as shown in Fig.~\ref{fig:LC}, the robot translates forward, rotates, and repeats -- continuously executing the limit cycle and counting how many times it bumps the object forward.

In the \emph{Left} substrategy, the robot orients itself so that it is facing the blue side of the object, then switches to \emph{Limit Cycle}, where it will translate toward the object, rotate, and enter the limit cycle.

In the \emph{Right} substrategy, $R_1$ will measure the distance to the object $y_{R\_old}$, move along the wall a small distance $\delta$, then measure the distance to the object again $y_R$, and compare the two distances. If the distance to the object increased, then the robot is moving toward the right and must turn around, using primitive $P_A$. Otherwise, the robot is moving toward the left, and will continue in that direction until it detects the blue side of the object and switches to the \emph{Left} substrategy.

In the \emph{Middle} substrategy, the robot is directly beneath or above the object, and cannot tell which direction is which. It chooses a random direction to follow the wall, until it detects either the blue or the yellow side of the object. If it detects blue it switches to the \emph{Left} substrategy, and if it detects yellow it switches to the \emph{Right} substrategy.

\begin{algorithm}
\caption{Robot 1 Policy: $\pi_1$} \label{alg:r1}

\begin{algorithmic}
\STATE \textbf{Requires} 
\STATE Primitives: $P_A$, $P_T$, $P_B$, $P_R$, $P_L$, $P_Y$
\STATE Parameters: $w$ is the range of distances that are attracted by the limit cycle
\STATE \textcolor{white}{.}
 
\STATE \textbf{Initial}
\STATE $count = 0$
\STATE $y_R, y_B, y_Y =$ observe\_object() \hspace*{\fill}(read object distance and color)
\IF{$y_B=1$ and $y_R \in w$ \hspace*{\fill}(if blue was detected at a distance in $w$)}
    \STATE Switch to \textbf{Limit Cycle}
\ELSIF{$y_B=1$ and $y_R \notin w$ \hspace*{\fill}(if blue was detected at a distance \emph{not} in $w$)}
    \STATE Switch to \textbf{Left}
\ELSIF{$y_Y=1$ \hspace*{\fill}(if yellow was detected)}
    \STATE Switch to \textbf{Right}
\ELSE{}
    \STATE Switch to \textbf{Middle}
\ENDIF

\STATE \textcolor{white}{.}
\STATE \textcolor{white}{.}

\STATE \textbf{Limit Cycle}
\STATE $P_T$ \hspace*{\fill}(translate forward to an obstacle)
\STATE $P_A(u_A=\theta)$ \hspace*{\fill}(rotate $\theta$)
\STATE $P_T$ \hspace*{\fill}(translate forward to an obstacle)
\STATE $P_A(u_A=\theta)$ \hspace*{\fill}(rotate $\theta$)
\STATE $P_T$ \hspace*{\fill}(translate forward to an obstacle)
\STATE $P_A(u_A=\theta)$ \hspace*{\fill}(rotate $\theta$)
\STATE $count ++$

\STATE \textcolor{white}{.}
\STATE \textcolor{white}{.}

\STATE \textbf{Left} (bounce off of blue side of object)
\STATE aim\_toward\_blue()
\STATE Switch to \textbf{Limit Cycle}

\STATE \textcolor{white}{.}
\STATE \textcolor{white}{.}

\STATE \textbf{Right} (wall follow toward object)
\STATE $y_{R\_old} =$ observe\_object() \hspace*{\fill}(read object distance)
\STATE wall\_follow($u_L=\delta$) \hspace*{\fill}(step forward a small distance $\delta$)
\STATE $y_R =$ observe\_object()  \hspace*{\fill}(read object distance)
\IF{$y_R > y_{R\_old}$ \hspace*{\fill}(if the distance to the object increased)}
    \STATE $P_A(u_A=180^{\circ})$ \hspace*{\fill}(turn around)
    \ENDIF
\WHILE{$y_B = 0$ \hspace*{\fill}(while blue has not been detected)}
    \STATE wall\_follow($u_L=\delta$) \hspace*{\fill}(step forward a small distance $\delta$)
    \STATE $y_B =$ observe\_object()  \hspace*{\fill}(scan for object and record color)
    \ENDWHILE
\STATE Switch to \textbf{Left}

\STATE \textcolor{white}{.}
\STATE \textcolor{white}{.}

\STATE \textbf{Middle} (wall follow until blue or yellow detected)
\WHILE{$y_B=0$ and $y_Y=0$ \hspace*{\fill}(while neither blue nor}
\STATE \textcolor{white}{.}\hspace*{\fill}yellow have been detected)
\STATE wall\_follow($u_L=\delta$) \hspace*{\fill}(step forward a small distance $\delta$)
\STATE $y_B,y_Y =$ observe\_object()  \hspace*{\fill}(scan for object and check if blue)
\ENDWHILE
\IF{$y_B = 1$  \hspace*{\fill}(if blue has been detected)}
    \STATE Switch to \textbf{Left}
\ELSIF{$y_Y = 1$  \hspace*{\fill}(if yellow has been detected)}
    \STATE Switch to \textbf{Right}
    \ENDIF
    
\end{algorithmic}
\end{algorithm}

\newpage
\subsection*{Robot 2 Policy}

Robot $2$ uses the primitives $P_A$, $P_T$, $P_B$, and $P_R$, and requires knowledge of the set of distances $w$ from the object that allow the robot to fall into the limit cycle. First, in its initial state, $R_2$ measures the distance between it and the object, and the color directly in front of it (if any). It uses this knowledge to switch to a substrategy: either \emph{Limit Cycle}, \emph{Left}, or \emph{Lost}.

In the \emph{Limit Cycle} substrategy, as shown in Fig.~\ref{fig:LC}, the robot translates forward, rotates, and repeats -- continuously executing the limit cycle and counting how many times it bumps the object forward.

In the \emph{Left} substrategy, the robot orients itself so that it is facing the blue side of the object, then switches to \emph{Limit Cycle}, where it will translate toward the object, rotate, and enter the limit cycle.

In the \emph{Lost} substrategy, $R_2$ translates forward to a wall or the object, and then stays still. It can never recover from this state.

\begin{algorithm}
\caption{Robot 2 Policy: $\pi_2$} \label{alg:r2}
\begin{algorithmic}
\STATE \textbf{Requires} 
\STATE Primitives: $P_A$, $P_T$, $P_B$, $P_R$
\STATE Parameters: $w$ is the range of distances that are attracted by the limit cycle
\STATE \textcolor{white}{.}

\STATE \textbf{Initial}
\STATE $count = 0$
\STATE $y_R, y_B =$ observe\_object()  \hspace*{\fill}(read object distance and color)
\IF{$y_B=1$ and $y_R \in w$ \hspace*{\fill}(if blue was detected at a distance in $w$)}
    \STATE Switch to \textbf{Limit Cycle}
\ELSIF{$y_B=1$ and $y_R \notin w$ \hspace*{\fill}(if blue was detected at a distance \emph{not} in $w$)}
    \STATE Switch to \textbf{Left}
\ELSE{}
    \STATE Switch to \textbf{Lost}
\ENDIF

\STATE \textcolor{white}{.}
\STATE \textcolor{white}{.}

\STATE \textbf{Limit Cycle}
\STATE $P_T$ \hspace*{\fill}(translate forward to an obstacle)
\STATE $P_A(u_A=\theta)$ \hspace*{\fill}(rotate $\theta$)
\STATE $P_T$ \hspace*{\fill}(translate forward to an obstacle)
\STATE $P_A(u_A=\theta)$ \hspace*{\fill}(rotate $\theta$)
\STATE $P_T$ \hspace*{\fill}(translate forward to an obstacle)
\STATE $P_A(u_A=\theta)$ \hspace*{\fill}(rotate $\theta$)
\STATE $count ++$

\STATE \textcolor{white}{.}
\STATE \textcolor{white}{.}

\STATE \textbf{Left} (bounce off of blue side of object)
\STATE aim\_toward\_blue()
\STATE Switch to \textbf{Limit Cycle}

\STATE \textcolor{white}{.}
\STATE \textcolor{white}{.}

\STATE \textbf{Lost}
\STATE $P_T$ \hspace*{\fill}(translate forward to an obstacle)


\end{algorithmic}
\end{algorithm}

\newpage
\subsection*{Robot 3 Policy}

Robot $3$ uses the primitives $P_A$, $P_T$, and $P_B$. In its initial state, $R_3$ attempts to execute the limit cycle (repeating $P_T$ and $P_A$) six times -- which, if $R_3$ started in the limit cycle, would translate to two cycles and the robot would detect blue twice during those two cycles (each time it bumped the object). If this is the case, the robot increases its $count$ to $2$ and switches to the \emph{Limit Cycle} substrategy. If the robot attempts to execute two limit cycles and does \emph{not} detect blue exactly twice, that means it did not start with the correct initial conditions, and switches to the \emph{Lost} substrategy.

In the \emph{Limit Cycle} substrategy, as shown in Fig.~\ref{fig:LC}, the robot translates forward, rotates, and repeats -- continuously executing the limit cycle and counting how many times it bumps the object forward.

In the \emph{Lost} substrategy, $R_3$ translates forward to a wall or the object, and then stays still. It can never recover from this state.

\begin{algorithm}
\caption{Robot 3 Policy: $\pi_3$} \label{alg:r3}
\begin{algorithmic}
\STATE \textbf{Requires} 
\STATE Primitives: $P_A$, $P_T$, $P_B$
\STATE \textcolor{white}{.}

\STATE \textbf{Initial}
\STATE $count = 0$
\STATE $inc = 0$ \hspace*{\fill}(variable for counting bounces)
\STATE $B = 0$ \hspace*{\fill}(variable for counting instances of blue detection)
\WHILE{$B < 2$ and $inc < 6$ \hspace*{\fill}(while blue has been detected less than twice}
\STATE \textcolor{white}{.} \hspace*{\fill}and fewer than six bounces have been attempted)

    \STATE $inc ++$
    \IF{$P_B(y_B=1)$ \hspace*{\fill}(if blue has been detected)}
        \STATE $B ++$ \hspace*{\fill}(count one blue detection)
        \ENDIF
    \STATE $P_T$ \hspace*{\fill}(translate forward to an obstacle)
    \STATE $P_A(u_A=\theta)$ \hspace*{\fill}(rotate $\theta$)
\ENDWHILE
\IF{$B=2$ \hspace*{\fill}(if blue has been detected twice)}
    \STATE $count=2$
    \STATE Switch to \textbf{Limit Cycle}
\ELSIF{$inc \geq 6$ \hspace*{\fill}(if six bounces have been attempted)}
    \STATE Switch to \textbf{Lost}
\ENDIF

\STATE \textcolor{white}{.}
\STATE \textcolor{white}{.}

\STATE \textbf{Limit Cycle}
\STATE $P_T$ \hspace*{\fill}(translate forward to an obstacle)
\STATE $P_A(u_A=\theta)$ \hspace*{\fill}(rotate $\theta$)
\STATE $P_T$ \hspace*{\fill}(translate forward to an obstacle)
\STATE $P_A(u_A=\theta)$ \hspace*{\fill}(rotate $\theta$)
\STATE $P_T$ \hspace*{\fill}(translate forward to an obstacle)
\STATE $P_A(u_A=\theta)$ \hspace*{\fill}(rotate $\theta$)
\STATE $count ++$

\STATE \textcolor{white}{.}
\STATE \textcolor{white}{.}

\STATE \textbf{Lost}
\STATE $P_T$ \hspace*{\fill}(translate forward to an obstacle)

\end{algorithmic}
\end{algorithm}

\end{document}